\pdfoutput=1
\documentclass{article}
\usepackage{qtBench_arxiv_version,times, tikz}
\usetikzlibrary{arrows.meta}
\usepackage{listings}
\usepackage{amsmath,amsfonts,bm}

\def\eqref#1{equation~\ref{#1}}

\def\1{\bm{1}}

\DeclareMathAlphabet{\mathsfit}{\encodingdefault}{\sfdefault}{m}{sl}
\SetMathAlphabet{\mathsfit}{bold}{\encodingdefault}{\sfdefault}{bx}{n}

\usepackage{hyperref}
\usepackage{url}

\newcommand{\qstat}{\mathrm{qstat}}
\newcommand{\tstat}{\mathrm{tstat}}
\newcommand{\area}{\mathrm{area}}
\newcommand{\bounce}{\mathrm{bounce}}
\newcommand{\dyck}{\mathrm{Dyck}_n}

\usepackage{booktabs,pifont}
\providecommand{\cmark}{\ding{51}}
\providecommand{\xmark}{\ding{55}}
\providecommand{\cmark}{\ding{51}}
\providecommand{\xmark}{\ding{55}}

\title{AI Discovery in Combinatorics:\\ From Problems to Solutions \\and Back to Bigger Problems}

\author{Eugenio Cainelli \\
University of Bologna \\
\texttt{eugenio.cainelli2@unibo.it}
\And
Lorenzo Luccioli \\
University of Bologna \\
\texttt{lorenzo.luccioli2@unibo.it}
\AND
Alessandro Iraci \\
Pegaso University \\
\texttt{alessandro.iraci@unipegaso.it}
\And
Giovanni Paolini \\
University of Bologna \\
\texttt{g.paolini@unibo.it}
}

\iclrfinalcopy
\begin{document}

\maketitle
\lhead{Preprint}

\begin{abstract}
We introduce $qt$Bench, a benchmark of 29 open problems in algebraic combinatorics. These problems ask for a simple combinatorial algorithm that explains given polynomial coefficients or identities. To help AI agents search for such algorithms, we provide a scaffold for playing with combinatorial objects, countering cheating attempts, and testing candidate solutions. To explore how far frontier AI systems can go on these problems, we collaborated with OpenAI, whose researchers obtained proposed solutions to several of them. This highlights bigger meta-problems: is AI discovery at scale ethical? How should these results be released? How should proof digestion be organized? We discuss these questions and give our views.
\end{abstract}

\section{Introduction}
A recurring problem in science and mathematics is to induct a simple deterministic rule that fully explains observable data. The motion of celestial bodies is a famous example: patterns in planets' trajectories led Newton to a compact law of gravitation, capable of explaining a much richer collection of observations. Mathematics has its own instances of this class of problems. Consider the problem of counting labelled trees on \(n\) vertices. Direct enumeration for \(n=2,3,4,5,6,\ldots\) produces the seemingly unstructured sequence
$
1,\ 3,\ 16,\ 125,\ 1296,\ldots
$
yet Cayley's formula reveals that these numbers are exactly
$
n^{n-2}.
$
A collection of observations is thus explained perfectly by a simple rule. The focus of this paper is on mathematical problems of this kind, where approximating or fitting the observed data is simply wrong: a solution must discover a \emph{simple} deterministic algorithm that explains the observations \emph{exactly}.

As large language models (LLMs) and LLM-based agents become increasingly capable at mathematical and scientific reasoning, it is important to understand to which degree they can perform this kind of discovery. Frontier systems are now able to make meaningful progress on open research questions: recent successes include improving state-of-the-art bounds \citep{georgiev2025discovery}, extending or completing arguments whose main structure is already present in the literature \citep{diez2025malliavin}, and discovering counterexamples to long-standing conjectures \citep{alon2026unitdistance}. In several such settings, progress can be incremental: a better bound is already a contribution even if it is not optimal, and improvement to one step of an existing argument can be mathematically valuable on its own. The problems we consider here are different. The AI system is given data generated by an unknown deterministic rule and must induct the rule itself. There is no incremental contribution: a rule either reproduces the data exactly or it does not. Success may require forming hypotheses, testing them computationally, recognizing when an apparently promising direction contains no path to an exact solution, abandoning it, and stepping back to tackle the problem from a different perspective. We seek to understand how current models behave in this regime, which simultaneously demands \emph{long-horizon reasoning} and \emph{creative exploration}, while being able to \emph{verify} proposed solutions efficiently.

Problems from algebraic combinatorics, and in particular from $q,t$-combinatorics, provide an unusually clean setting in which to study these capabilities. Many such problems arise from a recurring phenomenon in algebra and representation theory: one encounters an infinite family of polynomials indexed by natural numbers, say $n \in \mathbb{N}$, defined in an algebraic or representation-theoretic way. When these polynomials are computed explicitly for small values of $n$ (e.g., $n\leq 7$), a clear pattern emerges: their coefficients are nonnegative integers, for reasons that are not immediately apparent from the original definition. Then two natural questions arise: are the coefficients natural for all $n\in\mathbb{N}$, and, if so, what combinatorial objects do these coefficients count? A combinatorial interpretation answers both questions at once: it explains why the coefficients are nonnegative integers and gives them a concrete meaning. Discovering such interpretations, and finding combinatorial explanations for identities among the resulting polynomials, is the focus of the problems considered in this paper.

With these classes of problems, we present $qt$Bench, a benchmark of 29 open problems and 1 solved calibration problem in $q,t$-combinatorics. Our framework is defined by four key contributions:

\begin{itemize}
    \item
    We present an open-source benchmark of 30 problems across $q$-statistic discovery, $t$-statistic discovery, and bijection discovery. Because the solutions are unknown (except for the single calibration problem)\footnote{While preparing this paper, problems 7 and 17 were solved and their solutions published on arXiv \citep{interdonato2026theta,ding2026qanalogues}.}, they do not exist in any training corpus, and any correct solution produced by a model would therefore signal genuine reasoning ability and autonomous discovery.

    \item

    We provide an agent scaffold for attempting these problems: for each one, models receive a code implementation of the combinatorial objects together with the target polynomials for a range of parameters. We also develop an automatic evaluation pipeline that checks whether a proposed statistic or bijection satisfies the required combinatorial identities and detects several forms of cheating.
    \item We present a baseline evaluation of GPT-6 Astra (High), with two hours per problem using our inference scaffold, together with results from a collaboration with OpenAI on evaluating their internal models. These results include promising candidate solutions that, if verified, would constitute valuable contributions to algebraic combinatorics.    
    \item Motivated by these results, we discuss the ethical questions raised by AI-driven math discovery at scale and present our perspective on how such research should be conducted.

\end{itemize}

\section{Related Work}
\label{related-work}

\subsection{Benchmarks of Open Research Problems in Mathematics}

As AI's mathematical capabilities grow, benchmarks are needed
to measure what these systems can do and where they still fall short.
Recent benchmarks have begun to test them on open research problems.
HorizonMath \citep{wang2026horizonmath} contains mostly unsolved problems
from eight areas of mathematics, with code to check proposed closed-form
expressions, improvements to known bounds, and new constructions.
FrontierMath: Open Problems \citep{epoch2026openproblems} also focuses on
open problems whose proposed solutions can be checked computationally,
using a program written for each problem. Formal Conjectures
\citep{firsching2026formal} instead provides statements formalized in
Lean~4, including both open conjectures and solved problems, so that
proposed proofs can be checked by the proof assistant.
Recent AI-assisted work has also drawn attention to Thomas Bloom's
Erd\H{o}s Problems database \citep{bloom2026erdos},
which collects problems posed by Erd\H{o}s and records progress toward
their solution. Our benchmark contributes to this line of work by focusing on open
research problems, specifically in the field of algebraic combinatorics.

\subsection{AI for Discovering Combinatorial Statistics and Bijections}

Recently, researchers have begun using AI to search for combinatorial
statistics and bijections. \citet{huang2025blackbox} trained a Transformer
on pairs of Dyck paths and used its attention patterns to find a new
algorithm for the zeta map \citep{haglund2008diagonal}.
\citet{brown2025bijection} used OpenEvolve, an open-source system that
combines LLM-generated programs with evolutionary search, to search
for bijections in three problems involving Dyck paths. They recovered
one known bijection but did not solve the open area-bounce problem
they studied. \citet{cainelli2026mapping} used machine learning to
discover statistics for $q,t$-Narayana polynomials, leading to an explicit bijection proving the $q,t$-symmetry of $N_{n,3}(q,t)$. The studies above use either ad-hoc models or LLMs embedded in
an evolutionary search procedure. In the present work, we investigate whether general-purpose LLM agents can make progress when given tools for exploring combinatorial objects and testing candidate solutions, while leaving them free to choose their own search strategy.

\section{$q,t$-Combinatorics Setup}
\label{sec:qt-combinatorics}

We study problems that ask for combinatorial interpretations of
polynomial coefficients and identities. We explain the three resulting types of discovery problems through a single running
example. A \emph{Dyck path} of semilength $n$ is a lattice path from $(0,0)$ to
$(n,n)$, using north and east unit steps, that stays weakly above the main
diagonal. We write $\mathrm{Dyck}_n$ for the set of such paths.
These objects are counted by the Catalan number
$\mathrm{Cat}_n=\frac{1}{n+1}\binom{2n}{n}$.

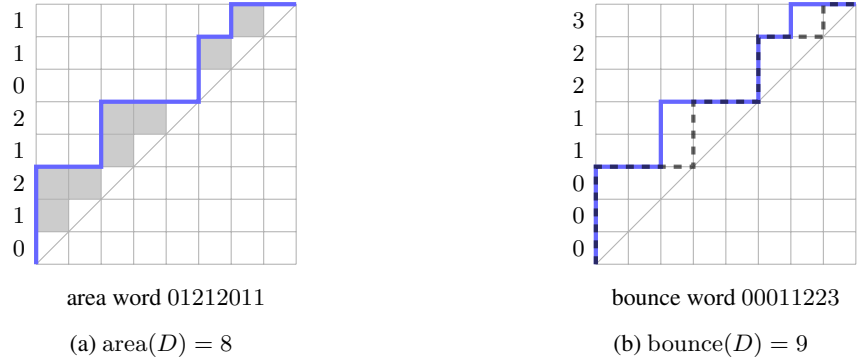
\begin{figure}[!ht]
    \centering
    \begin{minipage}[c]{0.47\linewidth}
        \centering
        \begin{tikzpicture}[scale=0.43]
            \foreach \x/\y in {0/1,0/2,1/2,2/3,2/4,3/4,5/6,6/7}
                \fill[gray,opacity=0.4] (\x,\y) rectangle ++(1,1);
            \draw[gray!60,thin] (0,0) grid (8,8);
            \draw[gray!60,thin] (0,0)--(8,8);
            \draw[blue!60,line width=1.6pt]
                (0,0)|-(2,3)|-(5,5)|-(6,7)|-(8,8);
            \foreach \y/\a in {0.5/0,1.5/1,2.5/2,3.5/1,4.5/2,5.5/0,6.5/1,7.5/1}
                \node[left] at (-0.05,\y) {\small $\a$};
            \node[below] at (4,-0.45) {\small area word $01212011$};
        \end{tikzpicture}
        \par\smallskip
        {\small (a) $\mathrm{area}(D)=8$}
    \end{minipage}
    \hfill
    \begin{minipage}[c]{0.47\linewidth}
        \centering
        \begin{tikzpicture}[scale=0.43]
            \draw[gray!60,thin] (0,0) grid (8,8);
            \draw[gray!60,thin] (0,0)--(8,8);
            \draw[blue!60,line width=1.6pt]
                (0,0)|-(2,3)|-(5,5)|-(6,7)|-(8,8);
            \draw[dashed,ultra thick,opacity=0.6]
                (0,0)|-(3,3)|-(5,5)|-(7,7)|-(8,8);
            \foreach \y/\b in {0.5/0,1.5/0,2.5/0,3.5/1,4.5/1,5.5/2,6.5/2,7.5/3}
                \node[left] at (-0.05,\y) {\small $\b$};
            \node[below] at (4,-0.45) {\small bounce word $00011223$};
        \end{tikzpicture}
        \par\smallskip
        {\small (b) $\mathrm{bounce}(D)=9$}
    \end{minipage}
    \caption{An $8\times 8$ Dyck path. In (a), each label on the left counts
    the shaded cells between the path and the main diagonal in that row.
    Read from bottom to top, these counts form the area word $01212011$,
    whose entries sum to $\mathrm{area}(D)=8$. In (b), the dashed line is
    the bounce path, and the labels on the left give its vertical-step
    labels, again read from bottom to top. They form the bounce word
    $00011223$, whose entries sum to $\mathrm{bounce}(D)=9$.}
    \label{fig:dyck-area-bounce}
\end{figure}

\subsection{$q$-statistic discovery}

A $q$-statistic discovery problem starts with an infinite family of
polynomials $F_n(q)$. As our running example, take $F_n(q)$ to be the
$q$-Catalan polynomials \citep{carlitz1964lattice},\footnote{These are not the $q$-analogs $\frac{1}{[n+1]_q}\genfrac{[}{]}{0pt}{}{2n}{n}_q$ of the formula $\mathrm{Cat}_n=\frac{1}{n+1}\binom{2n}{n}$.} which arise naturally in algebraic combinatorics. Setting $q=1$, we obtain
$F_n(1)=\mathrm{Cat}_n$, the ordinary Catalan number. Since Catalan numbers
count Dyck paths, this suggests that the full polynomial might refine the
ordinary count.

The task is to find a natural function (called statistic)
\[
    \mathrm{stat}:\dyck\longrightarrow\mathbb{N}
    \qquad\text{such that}\qquad
    F_n(q)=\sum_{D\in\dyck}q^{\mathrm{stat}(D)}
\]
for every $n$. If this equality holds, then the coefficient of $q^k$ has a
concrete meaning: it is exactly the number of paths $D$ for which
$\mathrm{stat}(D)=k$. Finding such a statistic therefore explains why every
coefficient of $F_n(q)$ is a nonnegative integer.

For the $q$-Catalan polynomials, one solution is the \emph{area} statistic. It
counts the whole lattice cells between the path and the main diagonal, as in
Figure~\ref{fig:dyck-area-bounce}, and satisfies
\[
    F_n(q)=\sum_{D\in\dyck}q^{\area(D)}.
\]
For example, $F_3(q)=1+2q+q^2+q^3$: among the five Dyck paths of
semilength $3$, one has $\area$ $0$, two have $\area$ $1$, one has $\area$ $2$, and one
has $\area$ $3$.

The word \emph{natural} is important. If the polynomial $F_n(q)$ has coefficients $a_0, a_1, \dots$, one could arbitrarily partition Dyck paths into groups of sizes $a_0, a_1, \dots$ and assign value $i$ to every path in the group of size $a_i$. The resulting statistic would reproduce the polynomial by construction, without explaining why the family of polynomials arises from these objects. The goal is a simple rule or algorithm that depends on the structure of each object and works uniformly for the whole infinite family. There may be several such rules and the task is to discover any of them.

\subsection{$t$-statistic discovery}

A $t$-statistic discovery problem is similar, but starts from an infinite
family of polynomials in two variables, $F_n(q,t)$. Part of the combinatorial
interpretation is already known: we are given a family of objects and a
statistic $\qstat$ satisfying
\[
    F_n(q,1)=\sum_{D\in\dyck} q^{\qstat(D)}.
\]
What is missing is a natural partner $\tstat$ such that
\[
    F_n(q,t)=\sum_{D\in\dyck}
    q^{\qstat(D)}t^{\tstat(D)}.
\]

The coefficient of $q^i t^j$ counts how many paths have $(\qstat, \tstat) = (i, j)$. Setting $q=1$ forgets the known statistic and tells us only how many paths must receive each value of the missing statistic. To match the full polynomial $F_n(q,t)$, these counts must also be correct within each group of paths having the same value of $\qstat$.

The $q$-Catalan polynomials have a two-variable refinement, the $q,t$-Catalan polynomials $C_n(q,t)$. 
This more general family recovers the one-variable polynomials when either variable is set to $1$:\footnote{Setting $t=1/q$ and normalizing gives instead the $q$-analogs of the formula for $\mathrm{Cat}_n$ mentioned above: $q^{\binom{n}{2}}C_n(q,1/q)=\frac{1}{[n+1]_q}\genfrac{[}{]}{0pt}{}{2n}{n}_q$ \citep{garsia1996remarkable}. So they, too, arise naturally from $C_n(q,t)$.}
\[
    C_n(q,1)=F_n(q)
    \qquad\text{and}\qquad
    C_n(1,t)=F_n(t).
\]

Therefore, we can take $\area$ as the known statistic and $C_n(q,t)$ as the target polynomial. 
The problem becomes to find a second statistic $\tstat$ such that
\[
    C_n(q,t)=\sum_{D\in\dyck}
    q^{\area(D)}t^{\tstat(D)}.
\]

One natural solution is Haglund's \emph{bounce} statistic \cite{haglund2003bounce}. Its discovery
took eight years of sustained community effort and was a major breakthrough.
To define $\bounce$, associate with a Dyck path $D$ a new lattice path, called the
\emph{bounce path}. Beginning at $(0,0)$, it moves north until it reaches the
initial point of an east step of $D$, and then it moves east until it returns to the
main diagonal. This procedure is repeated until the path reaches $(n,n)$.
Its vertical steps are labelled starting from $0$, with the label increasing
by $1$ after each return to the diagonal. The $\bounce$ statistic is the sum of
these labels. An example is given in Figure~\ref{fig:dyck-area-bounce}. A
different statistic, \emph{dinv}, is defined using certain pairs of steps, and
it gives another solution to the same problem. Thus $\bounce$ and
$\mathrm{dinv}$ are natural statistics defined in different ways, but both, together with $\area$, explain the
same $q,t$-Catalan polynomials.

\subsection{Bijection discovery}

A bijection discovery problem starts from a situation in which the pair of relevant
statistics are already known, but we do not understand why their joint distribution
has a specific property. For instance, the $q,t$-Catalan polynomial is symmetric:
\[
    C_n(q,t)=C_n(t,q).
\]
Since
$C_n(q,t)=\sum_D q^{\area(D)}t^{\bounce(D)}$, this symmetry implies
that, for every pair of nonnegative integers $i,j\in\mathbb{N}$, there are as many Dyck paths with statistics $\area(D)=i$ and $\bounce(D)=j$ as there are paths with statistics $\area(D)=j$ and $\bounce(D)=i$. The natural question is why.

To explain this combinatorially, we would like to find a bijection
$\phi:\dyck\rightarrow\dyck$ satisfying, for every path
$D$,
\begin{equation}\label{eq:exchange}
    \area(\phi(D))=\bounce(D),
    \qquad
    \bounce(\phi(D))=\area(D).
\end{equation}

Such a bijection gives a direct combinatorial proof of the $q,t$-symmetry: every object
contributing to the coefficient of $q^i t^j$ is sent to an object contributing to the coefficient of $q^j t^i$. Although the symmetry of $q,t$-Catalan polynomials is known by algebraic methods, no explicit bijection exchanging $\area$ and $\bounce$ is known, despite partial results \citetext{\citealp[Theorem~3.12]{haglund2008diagonal}; \citealp{lee2018combinatorial,han2020chain,han2022chain,ayyer2025areabounce,hawkes2026involution}}; finding one is an important, decades-old open problem with possible implications beyond combinatorics, and one of the
problems in $qt$Bench.

Again, an arbitrary bijection satisfying (\ref{eq:exchange}) is not a solution. Once symmetry is known, one can
order the paths with statistics $(i,j)$ and match them to paths with statistics
$(j,i)$. Such a construction assumes the symmetry it is supposed to explain
and gives no combinatorial insight. The goal instead is an explicit algorithm that explains the symmetry.

\section{Benchmark Design}
\label{sec:benchmark-design}

The goal of $qt$Bench is to discover conjectural statistics and bijections, without requiring proofs of correctness. This reflects a common practice in algebraic combinatorics: a natural algorithm supported by extensive numerical evidence is taken seriously as a conjecture and provides solid grounds for expecting the required identity to hold in general for the whole infinite family of polynomials. Finding such a conjectural function is a substantial mathematical contribution in its own right. Proving it correct is a separate problem that may require significant further effort.

\subsection{Design Principles}

$qt$Bench follows four design principles. First, every problem must belong to
one of the three discovery categories introduced in Section~3: $q$-statistic
discovery, $t$-statistic discovery, or bijection discovery. These three categories are broad enough to cover many important problems in algebraic combinatorics. Second, each problem must be mathematically meaningful. It is either stated as an open
problem in the literature or arises naturally from established combinatorial
objects, polynomials, or identities, and may already have been posed informally
within the community. Third, except for the solved calibration problem, every
problem must be open. Since no accepted solution is available, a successful
result provides evidence of an AI agent's capability for this kind of discovery
and could itself contribute to algebraic combinatorics. Fourth, we must be able to compute the target polynomials for a sufficiently wide range of parameter values to support the search for a solution. The first few cases may reveal too little structure for an agent to form and test meaningful hypotheses. Generating these data should take at most tens of minutes of CPU time on a machine with 64 GiB of RAM, while checking a proposed solution should take less than ten minutes.

\subsection{Curation Procedure}

We identify candidate problems through a manual review of the literature in
$q,t$-combinatorics and through discussions with experts in the field. For each
candidate, we identify the main references defining the relevant objects,
polynomials, or identities. We
then verify, to the best of our knowledge, that no solution of the problem is
known. We describe the relevant literature precisely. Finally, we implement
the relevant combinatorial objects and known statistics and compute the given
polynomials for a finite range.

$qt$Bench contains 30 problems: 29 open problems and 1 solved calibration
problem. They are organized by discovery category, as summarized in
Table~\ref{tab:problem-categories}.

\begin{table}[!ht]
    \centering
    \small
    \caption{The 30 problems of $qt$Bench by discovery category.}
    \label{tab:problem-categories}
    \renewcommand{\arraystretch}{1.18}
    \begin{tabular}{@{}p{0.20\linewidth}p{0.68\linewidth}r@{}}
        \hline
        \textbf{Category} & \textbf{Problem description} & \textbf{Count} \\
        \hline
        $q$-statistic discovery &
        {\raggedright Given a combinatorial object and a one-variable polynomial
        family, find a statistic that generates the polynomial.\par}
        & 8 \tabularnewline
        $t$-statistic discovery &
        {\raggedright Given a combinatorial object, one (sometimes none) known
        statistic, and a multivariate polynomial, find the missing statistic(s)
        that generates the polynomial together with the known statistic
        (if available).\par}
        & 13 \tabularnewline
        bijection discovery &
        {\raggedright Given a combinatorial object and two statistics, find a
        bijection on the set of combinatorial objects satisfying the required
        identity.\par}
        & 9 \tabularnewline
        \hline
    \end{tabular}
\end{table}

\subsection{Problem Package}

Each problem comes with a self-contained statement describing the combinatorial
objects, how they are represented in Python, the known statistics, and the
target polynomials or identity to be explained. We provide references to the
relevant literature and examples of the combinatorial objects, together with
their known statistic values. The agents can use these examples and the code
for the known statistics to explore the problem and test new ideas. Each
statement specifies which Python functions to submit and provides a command for
evaluating candidate solutions. Most statistic problems ask for
\texttt{statistic(object)}, while bijection problems ask for
\texttt{forward(object)} and \texttt{inverse(object)}. We describe the scaffold for the evaluation in the next section.

\section{Automated Evaluation}
\label{sec:checker}
\label{automated-evaluation}

\subsection{Numerical Verification}

A proposed statistic can be tested by evaluating it on the combinatorial
objects and checking whether it reproduces the target polynomials over the
finite range for which they have been computed. For example, for a
$t$-statistic discovery problem with known statistic $\mathrm{qstat}$ and
submitted statistic $\mathrm{tstat}$, the checker computes
\begin{equation}
    \widehat F_n(q,t)=\sum_{D\in\dyck}
    q^{\mathrm{qstat}(D)}t^{\mathrm{tstat}(D)}
\end{equation}
over the same range of $n$ as the given ground-truth polynomials $F_n(q,t)$, then checks
that $\widehat F_n(q,t)=F_n(q,t)$ for every $n$
in this range. 

For bijection discovery, we check that \texttt{forward} and \texttt{inverse}
return valid objects, are mutually inverse, and satisfy the identities
required by the problem on the objects of the scored range.

\subsection{Cheating Attempts}

Even if a submission passes numerical verification over the entire given range, this alone does not tell us whether it has overfit the computed examples or provides a meaningful solution to the problem. As discussed in Section~\ref{sec:qt-combinatorics}, arbitrary assignments generate the target polynomials or identities while assuming the very property a combinatorial interpretation should prove. We call such functions \emph{cheating attempts}.

In discussions with experts in the field, we found no formal
criterion for what makes a statistic or bijection a meaningful combinatorial
construction, or what should be instead classified as a cheating attempt. A proof formalized in Lean or a similar proof assistant could
certify the required identities for all $n \in \mathbb{N}$, but would not by itself
establish that the combinatorial problem is solved.
Determining whether the submission solves the problem requires mathematical
judgment. In practice, a mathematician can distinguish a natural statistic from an arbitrary assignment by
inspecting its definition, even without a general formal criterion.
In principle, stricter mechanical checks could reject all cheating attempts, but they could 
also exclude meaningful solutions. We therefore choose to build a checker
whose acceptance is a necessary but not sufficient condition for a solution,
and then use an LLM judge to assess
whether the construction is natural or a cheating attempt. Our
tests suggest that frontier models can make this distinction quickly and reliably (as described in the Appendix \ref{app:semantic-judge-evaluation}),
motivating this design choice. Even so, a submission that passes
both the checker and the LLM judge is only a \emph{candidate} solution, since
expert mathematical review is always required.

\subsection{Checker Design}

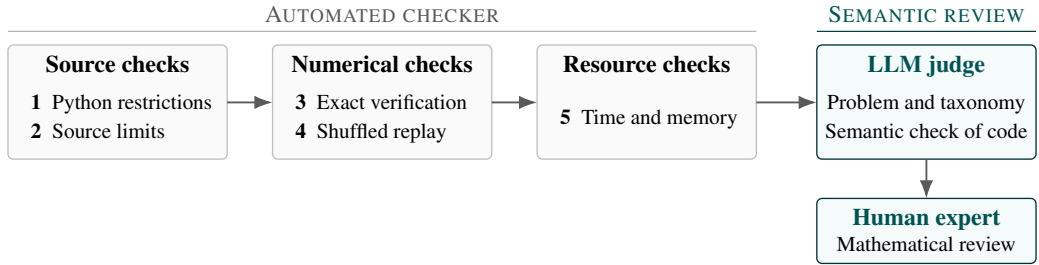
\begin{figure}[!ht]
    \centering
    \begin{tikzpicture}[
        x=1cm,y=1cm,
        every node/.style={font=\fontsize{8}{10}\selectfont},
        panel/.style={draw=black!25,fill=black!2,rounded corners=2pt,
                      line width=.5pt},
        flow/.style={-{Latex[length=2.4mm,width=1.8mm]},
                     line width=.7pt,draw=black!65},
        heading/.style={anchor=base,
                        font=\fontsize{9}{11}\selectfont\bfseries},
        note/.style={font=\scriptsize,text=black!65,align=center}
    ]
        \node[note,font=\footnotesize\scshape] at (4.95,1.02)
            {Automated checker};
        \draw[black!30,line width=.45pt] (0,.79) -- (9.90,.79);
        \node[note,font=\footnotesize\scshape,text=teal!65!black]
            at (12.15,1.02) {Semantic review};
        \draw[teal!45!black,line width=.45pt]
            (10.70,.79) -- (13.60,.79);

        \foreach \x in {0,3.5,7.0}
            \draw[panel] (\x,-.90) rectangle ++(2.90,1.51);

        \node[heading] at (1.45,.25) {Source checks};
        \node[anchor=base west] at (.18,-.24)
            {\textbf{1}~~Python restrictions};
        \node[anchor=base west] at (.18,-.62)
            {\textbf{2}~~Source limits};

        \node[heading] at (4.95,.25) {Numerical checks};
        \node[anchor=base west] at (3.68,-.24)
            {\textbf{3}~~Exact verification};
        \node[anchor=base west] at (3.68,-.62)
            {\textbf{4}~~Shuffled replay};

        \node[heading] at (8.45,.25) {Resource checks};
        \node[anchor=base west] at (7.18,-.43)
            {\textbf{5}~~Time and memory};

        \draw[panel,draw=teal!40!black,fill=teal!4]
            (10.70,-.90) rectangle (13.60,.61);
        \node[heading,text=teal!65!black] at (12.15,.25)
            {LLM judge};
        \node[anchor=base] at (12.15,-.24)
            {Problem and taxonomy};
        \node[anchor=base] at (12.15,-.62)
            {Semantic check of code};

        \draw[flow] (2.90,-.15) -- (3.50,-.15);
        \draw[flow] (6.40,-.15) -- (7.00,-.15);
        \draw[flow] (9.90,-.15) -- (10.70,-.15);
        \draw[flow] (12.15,-.90) -- (12.15,-1.42);

        \draw[panel,draw=teal!40!black,fill=teal!4]
            (10.70,-2.28) rectangle (13.60,-1.42);
        \node[heading,text=teal!65!black] at (12.15,-1.77)
            {Human expert};
        \node[anchor=base] at (12.15,-2.12)
            {Mathematical review};
    \end{tikzpicture}
    \caption{Evaluation pipeline for benchmark problems.}
    \label{fig:checker-pipeline}
\end{figure}

We ask for short, self-contained Python submissions and evaluate them using the same five checker stages for every problem. The stages run in order, and evaluation stops at the first failure. Submissions that pass all five stages proceed to the LLM judge, which assesses whether they provide a meaningful combinatorial construction. Those approved by the judge should still be reviewed by an expert mathematician (Figure~\ref{fig:checker-pipeline}).

\paragraph{1 and 2. Source checks.}
We first inspect the submitted code without running it. For both
statistic and bijection discovery, we require a single, self-contained
Python file. Submissions may define helper functions and use loops,
recursion, and built-in data structures. They may not import modules
or define classes, and they may use only a small set of Python's
built-in functions. The first check, a \emph{capability screen},
enforces these rules and also rejects code that accesses files, uses
reflection, or reaches into Python's runtime internals.

The second check limits the size of the source code. Since we expect
statistics and bijections to admit short implementations, we
apply the same limits to both task types: at most 8,192 bytes,
160 lines, 1,500 tokens (excluding comments and whitespace), 2,000
nodes in the abstract syntax tree, and 256 bytes of literal constants
in total. These limits are intended to exclude submissions that store
precomputed answers instead of implementing a combinatorial rule, and
to keep submissions easy to review. 

\paragraph{3 and 4. Numerical checks.}
We then run the submission on the finite collection of combinatorial
objects selected for the problem. For statistics, we check that the
resulting polynomials equal the supplied target polynomials.
For bijections, we check that the outputs are valid objects,
that the composition of the forward and inverse maps is equal to the identity map, and that the required identities hold. 

Matching the target polynomials does not ensure that the submitted
function assigns a deterministic value to each object: a program could
distribute values according to the order in which objects are received.
We therefore reload passing submissions and evaluate the same objects
in a randomly shuffled order, with repeated calls. Each object must
produce the same output as before. This detects dependence on call
order or previous evaluations, but does not exclude arbitrary
assignments that consistently associate the same output with each object.

\paragraph{5. Resource checks.}
We use time and memory limits to reject submissions that are too
computationally expensive to evaluate. The numerical checks must
finish within 60 seconds for statistics and 480 seconds for
bijections.

\subsection{Semantic Review and Acceptance}

We maintain a taxonomy of cheating attempts that describes how they
work, which checks detect them, and which attempts can still pass the checker.
To expand this taxonomy, we conducted adversarial experiments in which
LLMs were asked to produce submissions that pass the checker without
solving the underlying combinatorial problem. These experiments helped
us identify more cheating strategies and document the limitations
of the checker.

The LLM judge, running in an isolated sandbox, receives the taxonomy together with the problem statement
and the submitted code. It uses this context to assess whether the
submission provides a meaningful combinatorial construction or constitutes
a cheating attempt. The taxonomy informs this assessment but does not
exhaust the possible ways to cheat. In our tests, the LLM judge distinguishes meaningful proposals from cheating attempts with high accuracy; we present the evaluation
and its results in the Appendix \ref{app:semantic-judge-evaluation}.

A submission is accepted as a candidate only if it passes all checker
stages and receives the LLM judge's approval. 

Getting agents to find genuine solutions  is not only a mathematical
problem, but also an alignment problem. When the models cheat, it is
neither for lack of context, since the statements and our instructions
spell out what counts as a solution and what counts as cheating, nor
for lack of understanding, since frontier models reliably make this
distinction when we use them as judges. They cheat because they are
not well aligned. This applies to both GPT-6 Astra (High) and OpenAI's
internal models: cheating was common in our baseline runs, and at
least 4 of the 21 solutions proposed by OpenAI's models rely on some
form of cheating. In our baseline evaluation, the reasoning traces exhibit a clear shift: it happens frequently that once a problem is deemed too hard, the model stops
working on it and tries to game the verifier.

\section{AI Discovery at Scale, Responsible Release and Community Effort}
\label{sec:responsible-release}

We created this benchmark because we thought that publicly available models
would be unable to solve any of these open problems, with their tendency
toward ``low-entropy'' thinking being the main capability bottleneck.
In fact, in our experiments, which consisted in two hours of inference on each
problem using GPT-6 Astra (High), no proposed solution has passed both the checker and the LLM judge.

In July 2026, we also began a collaboration with OpenAI to test their internal models and inference scaffold on these problems. We did not expect notable results, but OpenAI returned proposed solutions to 21 problems in the benchmark,\footnote{The proposed solution to Problem~8 is part of OpenAI's major release of mathematical results in October 2026 \citep{OAI:Elementary-Positivity-of-Chromatic-Quasisymmetric-Functions-September-24-2026}.} 10 of which we consider genuine solutions to the given problems (see Appendix~\ref{app:openai-proposals}). These include a bijective proof of the \(m,n\)-symmetry of the \(q,t\)-Narayana polynomials \citep{aval2014statistics} and counterexamples to 3 positivity conjectures. We also received a proposed bijection for the symmetry of the \(q,t\)-Catalan polynomials, guided by a known algebraic proof \citep{carlsson2018proof}, which we still need to assess.

The LLM-generated material is poorly written and
requires substantial work to understand and rewrite the solutions.
This work is akin to what \citet{tao2026mathematics} calls \emph{proof digestion}.
Given the importance of these problems to algebraic
combinatorics, and the increasing capabilities of LLMs in 
mathematics, we would like to open a discussion around the following questions,
and give our own views in the rest of this section:

\begin{enumerate}
\item Is it ethical to use AI to solve a substantial proportion of a field's open problems?
\item How should the results be released? 
\item How should the work of proof digestion be organized?
\item How should credit be allocated?
\end{enumerate}

\subsection{Ethics of AI-Assisted Discovery}
\label{sec:discovery-ethics}

What is mathematics, and what is its goal? Neither question has an easy answer. Yet, in these turbulent times, much of the mathematical community seems to be returning to the view expressed by William Thurston in \emph{On proof and progress in mathematics} \citep{thurston1994proof}: mathematical progress is about advancing human understanding, not just producing proofs. We take clarity and understanding to be a primary goal of mathematics throughout this section. With this goal in mind, is it good to use AI to prove results in a field at scale?

We believe that the new information supplied by AI in the form of proofs and counterexamples is neither inherently good nor inherently bad. Its value depends on how the community works with it and on the incentives surrounding that work. Faced with AI discovery at scale, we believe there is a set of trajectories of choices the community can follow that would have a net positive impact on the clarity and understanding of a field. Does leaving its traditions and incentive system unchanged belong to any of these trajectories? We do not think so. The community needs to change radically to respond to the increasing capabilities of AI. For the changes we support, we refer to Daniel Litt's essay \emph{A beginning for mathematics} \citep{litt2026beginning} and to Timothy Gowers's recent blog post \citep{gowers2026fields}, which largely reflect our views on policy changes for the mathematical community.

The community cannot avoid these questions simply by calling for mathematicians to stop using AI. We expect tools with capabilities comparable to those now available internally at OpenAI to become widely accessible in the near future, as much as computers are available today. Once these tools are widely available, it would be unrealistic to enforce a blanket ban on their use, even within academia. Hence, the community needs to decide how to engage with AI-generated discoveries and turn them into shared mathematical understanding.

OpenAI themselves started an independent advisory group \citep{openai2026release} to discuss these important questions; we look forward to seeing the outcome.

\subsection{Release of Results}
\label{sec:results-release}

We believe it is fair to give the community access to AI-generated proposed solutions, since we are not able to digest all of them. At the same time, we do not want to flood arXiv, or the wider mathematical discussion, with poorly written AI-generated papers. We therefore propose to share them with researchers in the field that are interested in the digestion work, in writing a paper on the material they are able to digest, and in disseminating their understanding through talks at conferences. 

We cannot control how these documents are used after their release. We therefore advocate for the community to digest the solutions before building on the results. Skipping this step risks piling poorly understood papers on top of one another, taking us further from the clarity and understanding we seek to achieve. 

\subsection{Proof Digestion}
\label{sec:proof-digestion}

Part of the backlash against AI-generated solutions to important mathematical problems concerns the failure to properly acknowledge earlier work \citep[see, e.g.,][]{binder2026leiden}. When insights developed by mathematicians are presented without due credit, we believe the criticism is justified. An important part of proof digestion is therefore to establish which ideas are new and which are already present in the literature, even when the AI-generated text does not acknowledge them.

Once the LLM-generated documents are released, we suggest that groups interested in digesting a particular result discuss their plans openly, much as the Mathlib community coordinates work on Zulip. When a group has completed this work, its members can publish a paper presenting the digested result, stating in the acknowledgments or AI disclosure that their work builds on an AI-generated discovery. New results can then build on the digested proofs, with and without AI.

Should AI-generated results also be presented at conferences and workshops, as human-generated results are? We believe they should. Sharing new mathematics is essential to the progress of the field, regardless of whether AI contributed to its discovery. Worthy results, ideas, and techniques need to be understood by the community and become part of its mathematical toolkit. We believe that the people who digested the solutions and wrote the resulting papers are best placed to present this work and have earned that role through their efforts.

\subsection{Credit and Attribution}
\label{sec:credit-attribution}

We cannot dictate how credit should be allocated, but we highlight three main types of contribution applicable to the $qt$Bench story: \textbf{problem selection, scaffold building, and inference},
the contribution described in this paper; \textbf{AI training}, the work of the company that trained the models; and \textbf{digestion and
dissemination}, the work of those who digest the solutions, write the
resulting papers, and explain the results to the community.

We believe the mathematical community should place increasing importance
on the first and third contributions: choosing questions worth pursuing
and turning their answers into shared understanding. Problem
solving might never be fully automated, but advances in AI may soon decrease its
relative importance in mathematical research. However, even AI that can
discover, prove, and explain new mathematics does not remove the need
for people to understand it. The future of mathematics depends not only
on what machines can discover, but on what we humans come to understand.

\subsection*{AI use statement}
We used AI tools to help implement the benchmark and evaluation
code, generate and assess candidate solutions, create the synthetic
submissions used in Appendix~\ref{app:semantic-judge-evaluation}, and edit the paper. We manually
reviewed the synthetic submissions and their labels. We take full
responsibility for the paper's content.

\subsection*{Ethics statement}
Section~\ref{sec:responsible-release} discusses the ethics of our work.

\subsection*{Reproducibility statement}
The benchmark code and data are available at \url{https://github.com/SwagCai/qtBench}; installation and evaluation instructions are provided in \texttt{README.md}.

For Problem~3, the
code behind the target polynomials with $n \ge 5$ is not included, and the
targets for $n = 7, 8, 9$ are conditional on some conjectures on the type $B$ $q,t$-Catalan polynomials, not computed
directly. Our baseline evaluation uses a publicly available model, GPT-6 Astra (High),
although reruns may give different outputs; OpenAI's internal runs are not reproducible.

\subsection*{Acknowledgments}
We thank Omar Ismail and Sang Michael Xie, members of technical staff at OpenAI, for the collaboration on $qt$Bench.
We also thank other members of technical staff at OpenAI for private discussions on $qt$Bench.
The authors are grateful to the organizers of several summer schools and conferences where fruitful discussions on this work took place, including the summer school \textit{Geometry, Algebra and Combinatorics of Moduli Spaces and Configurations} in Dobbiaco (2026), the C.I.M.E. summer school \textit{Algebraic and geometric combinatorics: Coxeter group theory and representation theory} in Cetraro (2026), and \textit{ICML 2026}.

\bibliography{qtBench_arxiv_version}
\bibliographystyle{qtBench_arxiv_version}

\appendix
\section{Evaluation of the Semantic Judge}
\label{app:semantic-judge-evaluation}

To test whether an LLM semantic judge could distinguish legitimate functions from cheating attempts, we prepared 51 synthetic problems on Dyck paths: 17 for $q$-statistic discovery, 17 for $t$-statistic discovery, and 17 for bijection discovery. We asked GPT-6 Astra to generate two submissions for each problem: one legitimate and one adversarial, designed to hide a cheating strategy that would be difficult for the judge to detect. We also instructed the model to make the submissions similar in several respects, such as code length and coding style.

We reviewed all 102 submissions and confirmed their intended labels. We then evaluated GPT-6 Astra (Medium) as a semantic judge on this synthetic dataset. Each submission was judged separately in a fresh context containing the problem statement, submitted code, and cheating taxonomy. The judge had no access to external tools and a time limit of 120 seconds per evaluation.

The judge correctly classified all 51 legitimate submissions and all 51 cheating attempts, completing every evaluation in under 14 seconds.

Although the problems were synthetic and the functions not the most complex, these results show that Astra could quickly recognize adversarial cheating attempts in this dataset. Its accuracy supports our use of an LLM judge informed by the cheating taxonomy in the $qt$Bench evaluation pipeline.

We illustrate the experiment below with one problem and its two submissions, one legitimate and one cheating.

\subsection{Example: boundary runs of Dyck paths}

\paragraph{Problem.}
Let $\mathcal{D}_n$ be the set of words with $n$ letters $N$ and $n$ letters $E$ such that every prefix contains at least as many $N$'s as $E$'s.\footnote{This is the definition of Dyck paths as in Section~\ref{sec:qt-combinatorics}.} For $D\in\mathcal{D}_n$, let $a(D)$ be the length of its initial run of $N$'s and $b(D)$ the length of its terminal run of $E$'s, with both equal to zero for the empty path. Construct a uniform, combinatorially meaningful bijection $\phi:\mathcal{D}_n\to\mathcal{D}_n$ satisfying
\[
a(\phi(D))=b(D),\qquad b(\phi(D))=a(D).
\]
Supply mutually inverse Python functions \texttt{forward(path)} and \texttt{inverse(path)}, and explain the map through the structure of paths. Such a map explains the symmetry of the polynomials $F_n(q,t)=\sum_{D\in\mathcal{D}_n}q^{a(D)}t^{b(D)}$.

\paragraph{Legitimate submission.}
The first submission interprets the path as the contour of a plane rooted tree and reverses the order of children at every vertex. This is equivalent to reversing the word and interchanging $N$ and $E$. The transformation preserves Dyck paths, exchanges the two boundary runs, and is an involution.

\begin{lstlisting}
def decode(path):
    root = []
    stack = [root]
    for c in path:
        if c == 'N':
            child = []
            stack[-1].append(child)
            stack.append(child)
        else:
            stack.pop()
    return root

def encode(tree):
    output = []
    pending = [('visit', tree)]
    while pending:
        kind, value = pending.pop()
        if kind == 'letter':
            output.append(value)
        else:
            for child in value:
                pending.append(('letter', 'E'))
                pending.append(('visit', child))
                pending.append(('letter', 'N'))
    return ''.join(output)

def forward(path):
    return encode(decode(path))

def inverse(path):
    return encode(decode(path))
\end{lstlisting}

\paragraph{Judge assessment (legitimate).}
The judge classified this submission as legitimate and gave the following explanation:
\begin{quote}
\texttt{decode} constructs the rooted ordered tree encoded by the Dyck path. In \texttt{encode}, the LIFO stack processes each node's children in reverse order, thus reflecting the tree at every node. This uniform structural involution exchanges the leftmost and rightmost branch lengths, which are the initial N-run and terminal E-run. Both functions implement this same involution; there is no target lookup, fiber ranking, enumeration, or finite-range fitting.
\end{quote}
\paragraph{Cheating submission.}
The second submission enumerates the paths with boundary statistics $(a,b)$ and $(b,a)$, then maps each path to the one with the same position in the other list. This produces a bijection because the two lists have equal size, but gives no structural explanation for that equality. It is therefore a rank-matching cheating attempt under the taxonomy, even though it satisfies the numerical identities.

\begin{lstlisting}
def boundaries(path):
    return (len(path) - len(path.lstrip('N')), len(path) - len(path.rstrip('E')))

def completions(n, first, last):
    entries = [(i, 'N') for i in range(first)] + [(first, 'E')]
    entries += [(2 * n - last - 1, 'N')]
    entries += [(i, 'E') for i in range(2 * n - last, 2 * n)]
    required = dict(entries)
    if any((required[i] != c for i, c in entries)):
        return []

    def extend(prefix, height):
        i = len(prefix)
        if height < 0 or height > 2 * n - i:
            return []
        if i == 2 * n:
            return [prefix]
        return [word for c, step in [('E', -1), ('N', 1)] if required.get(i, c) == c for word in extend(prefix + c, height + step)]
    return extend('', 0)

def forward(path):
    if not path:
        return ''
    x, y = boundaries(path)
    n = len(path) // 2
    source = completions(n, x, y)
    target = completions(n, y, x)
    return target[source.index(path)]

def inverse(path):
    return forward(path)
\end{lstlisting}

\paragraph{Judge assessment (cheating).}
The judge classified this submission as cheating and gave the following explanation:
\begin{quote}
\texttt{completions} exhaustively enumerates the Dyck paths with specified boundary runs in a fixed E-before-N order. \texttt{forward} builds the $(x,y)$ and $(y,x)$ fibers and returns \texttt{target[source.index(path)]}, pairing paths solely by their enumeration ranks rather than transforming path structure. This is a uniform rank-matched bijection implemented by exhaustive fiber enumeration, not a combinatorially meaningful construction.
\end{quote}

The judge accepted the first submission and flagged the second.

\section{Overview of Problems and OpenAI's Proposed Solutions}
\label{app:openai-proposals}

Table~\ref{tab:openai-proposals} lists all 30 problems of $qt$Bench with the status of the corresponding proposal in the OpenAI package or, for Problem~8, in OpenAI's October 2026 release \citep{OAI:Elementary-Positivity-of-Chromatic-Quasisymmetric-Functions-September-24-2026}. In total, there are 18 proposed constructions and 3 proposed counterexamples. The package includes code for 10 of the constructions and scripts for
the 3 counterexamples; the other 8 constructions are write-ups only.

The marks are our informal assessment, they do not certify correctness or novelty, and they are not benchmark scores. In particular, we assessed every proposal in the package, including any that the checker or the LLM judge rejected; the eight constructions without code could not be run through the checker at all. A tick (\cmark) marks a direct construction, which computes the statistic or bijection from the structure of each object,
as asked in Section~3, and the construction passes both the checker and the LLM judge assessment. A question mark (\textbf{?})\ marks proposed solutions that we are still assessing. These constructions are not implemented in code (except for problem 28), hence the checker did not validate or reject them. A cross (\xmark) marks a construction
that is clear cheating and of no mathematical value; such construction may pass the checker but not the LLM judge assessment. 

\begin{table}[h!]
\centering
\small
\setlength{\tabcolsep}{4pt}
\begin{tabular}{@{}r p{0.6\linewidth} c c c@{}}
\toprule
ID & Problem & Type & Code & Assessment \\
\midrule
1  & Noncrossing $q,t$-Narayana partner (calibration)         & $t$ &     & ---      \\
2  & Dyck paths: area/bounce exchange                         & B   & no  & \textbf{?} \\
3  & Type $B$ $q,t$-Catalan area partner                      & $t$ &     & ---      \\
4  & Polyominoes: area/bounce exchange                        & B   & no  & \textbf{?} \\
5  & Polyominoes: $(m,n)$-transpose bijection                 & B   & yes & \cmark   \\
6  & Labelled polyominoes: Theta-operator area partner        & $t$ & yes & \cmark   \\
7  & Doubly decorated Dyck paths: unified-Delta partner       & $t$ & no  & \xmark   \\
8  & Shareshian-Wachs $e$-positivity statistic                 & $q$ & no  & \textbf{?} \\
9  & Zero-rooted tiered trees: $\Xi$-operator partner         & $t$ & yes & \textbf{C}        \\
10 & Multi-labelled Dyck paths: super-nabla partner           & $t$ & yes & \textbf{C}        \\
11 & Rise-decorated rectangular paths: Delta partner          & $t$ & yes & \textbf{C}        \\
12 & Tamari Hilbert-series shadow: third statistic            & $t$ &     & ---      \\
13 & Modified $(q,t)$-Kostka tableau statistics               & $t$ &     & ---      \\
14 & Unicellular LLT Schur-coefficient statistic              & $q$ & no  & \xmark   \\
15 & Involutions: orbit-harmonics Hilbert-series statistic    & $q$ & yes & \cmark   \\
16 & Matchings: Jack connection-coefficient statistic         & $q$ &     & ---      \\
17 & Permutations: $q$-Eulerian gamma statistic               & $q$ & yes & \cmark   \\
18 & Selected-area gamma-parking: Delta-$\Xi$ partner        & $t$ &     & ---      \\
19 & Rooted tiered trees: Theta-operator partner              & $t$ &     & ---      \\
20 & Selected-area lattice gamma-parking Schur partner        & $t$ &     & ---      \\
21 & Threshold-graph trees: flow-polytope Ehrhart partner     & $t$ &     & ---      \\
22 & Promotion tableaux: intrinsic statistic                  & $q$ & yes & \cmark   \\
23 & Noncrossing partitions: $q$-Kreweras statistic           & $q$ & yes & \cmark   \\
24 & Macdonald fillings: $\mathrm{inv}/\mathrm{maj}$ exchange & B   & no  & \textbf{?} \\
25 & Parking functions: area/$\mathrm{dinv}$ exchange         & B   & no  & \textbf{?} \\
26 & Connected graphs: sibling/tuft exchange                  & B   & yes & \xmark   \\
27 & Alternating sign matrices: DPP-weight statistic          & $q$ & no  & \textbf{?} \\
28 & Shifted set-valued tableaux: $P/Q$ correspondence        & B   & yes & \textbf{?} \\
29 & Andrews-Bressoud successive-rank bijection              & B   & yes & \xmark   \\
30 & Partition matrices: inversion-sequence bijection         & B   & yes & \cmark   \\
\bottomrule
\end{tabular}
\caption{Status of the OpenAI proposals for the 30 $qt$Bench problems.
\emph{Type}: $q$-statistic ($q$), $t$-statistic ($t$) or bijection (B)
discovery, as in Table~1. \emph{Code}: whether the package includes code (for
counterexamples, a verification script). The marks are an informal
assessment.}
\label{tab:openai-proposals}
\end{table}

\paragraph{Counterexamples.}
The three counterexamples are about the target polynomials themselves, not about
candidate statistics. In each case, the target polynomial has a negative coefficient outside the
evaluated parameter range: for Problem~9, the coefficient of $q^1t^0$ at tier sizes $(10,9,9)$ is
$-2{,}871{,}205{,}236{,}900$; for Problem~10, the constant term at $n=7$,
$k=2$ is $-8{,}484$; and for Problem~11, the constant term at
$(m,n,k)=(2,11,1)$ is $-528$. The generating function of a statistic has
nonnegative coefficients, so no statistic can generate these polynomials. These counterexamples are of considerable value to algebraic combinatorics community.

\end{document}